\documentclass[letterpaper]{article} % DO NOT CHANGE THIS
\usepackage{aaai2026}  % DO NOT CHANGE THIS
\usepackage{times}  % DO NOT CHANGE THIS
\usepackage{helvet}  % DO NOT CHANGE THIS
\usepackage{courier}  % DO NOT CHANGE THIS
\usepackage[hyphens]{url}  % DO NOT CHANGE THIS
\usepackage{graphicx} % DO NOT CHANGE THIS
\def\UrlFont{\rm}  % DO NOT CHANGE THIS
\usepackage{natbib}  % DO NOT CHANGE THIS AND DO NOT ADD ANY OPTIONS TO IT
\usepackage{caption} % DO NOT CHANGE THIS AND DO NOT ADD ANY OPTIONS TO IT
\usepackage{algorithm}
\usepackage{algorithmic}

\usepackage{newfloat}
\usepackage{listings}
\DeclareCaptionStyle{ruled}{labelfont=normalfont,labelsep=colon,strut=off} % DO NOT CHANGE THIS
\floatstyle{ruled}
\newfloat{listing}{tb}{lst}{}
\floatname{listing}{Listing}
\usepackage{comment}

\title{ORACLE: Optimizing Reasoning Abilities of Large Language Models via Constraint-Led Synthetic Data Elicitation}
\author{
    Zhuojie Yang\textsuperscript{\rm 1}\equalcontrib, 
    Wentao Wan\textsuperscript{\rm 1}\equalcontrib, 
    Keze Wang\textsuperscript{\rm 1}\thanks{Corresponding author.}
}
\affiliations{
    \textsuperscript{\rm 1}School of Computer Science and Engineering, Sun Yat-sen University\\
    \{yangzhj53, wangkeze\}@mail2.sysu.edu.cn, wanwentao93@gmail.com
}

\usepackage{bibentry}
\begin{document}

\maketitle

\begin{abstract}
Training large language models (LLMs) with synthetic reasoning data has become a popular approach to enhancing their reasoning capabilities, while a key factor influencing the effectiveness of this paradigm is the quality of the generated multi-step reasoning data. To generate high-quality reasoning data, many recent methods generate synthetic reasoning paths and filter them based on final answer correctness, often overlooking flaws in intermediate reasoning steps. To enhance the verification of intermediate reasoning steps, prior work primarily resorts to code execution or symbolic reasoning engines. However, code-based validation is restricted to code or mathematical tasks, and reasoning engines require a well-structured and complete context. As a result, existing methods fail to function effectively in natural language reasoning tasks that involve ambiguous or incomplete contexts. In these tasks, synthetic data still lack reliable checks for verifying each reasoning step. To address this challenge, we introduce \textbf{ORACLE}, a structured data generation framework inspired by syllogistic reasoning. ORACLE integrates the generative strengths of LLMs with symbolic supervision: the LLM produces step-wise reasoning contexts, while a symbolic reasoning engine verifies the validity of each intermediate step. By employing a unified prompting template to elicit modular reasoning chains, ORACLE enables fine-grained, step-level validation, facilitating the construction of high-quality multi-step reasoning data. Across six logical, factual, and commonsense reasoning benchmarks, our ORACLE consistently outperforms strong baselines on multiple models.
\end{abstract}

\begin{links}
    \link{Code}{https://github.com/yangzhj53/ORACLE}
\end{links}

\section{Introduction}

\begin{figure*}[t]
  \centering
  \includegraphics[width=0.65\textwidth]{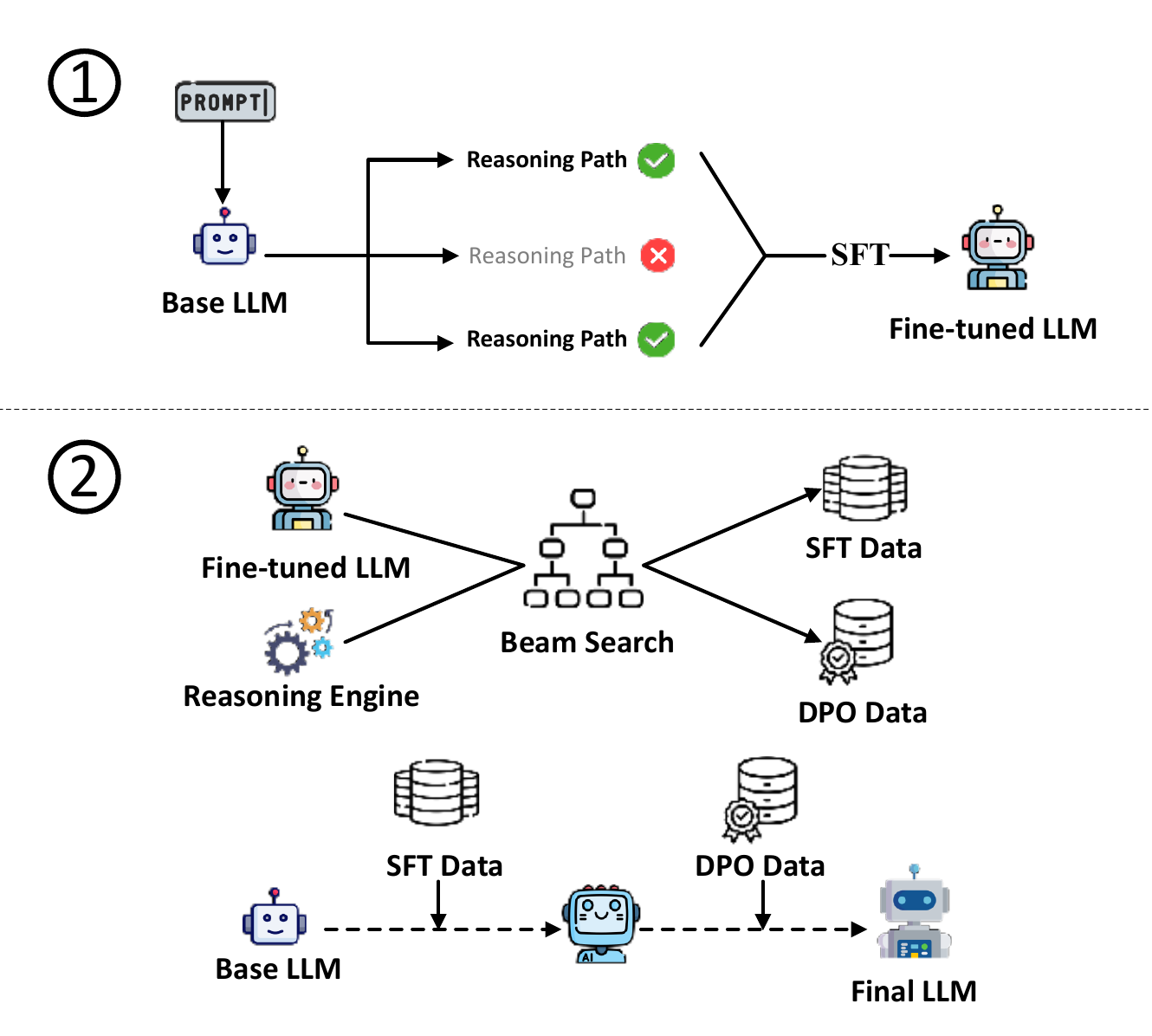}
  \caption{An overview of two-stage training pipeline of our ORACLE. ORACLE adopts a two-stage training pipeline: Stage 1 employs few-shot prompting and template-based reasoning generation, followed by answer-based and format-based filtering; Stage 2 integrates symbolic reasoning via beam search to produce high-quality reasoning data for supervised fine-tuning(SFT) and Direct Preference Optimization(DPO).}
  \label{fig:training_pipeline}
\end{figure*}

Training large language models (LLMs)~\cite{zhao2023survey,chang2024survey,hadi2023survey, liu2024deepseek} with synthetic reasoning data has emerged as a widely adopted paradigm for improving their performance on complex reasoning tasks such as multi-hop question answering, mathematical problem solving, and scientific inquiry\cite{zelikman2022star, yuan2023scaling, xu2023selfrewarding}. This approach leverages the generative capabilities of LLMs to scale up the creation of multi-step reasoning examples, which are otherwise expensive and labor-intensive to annotate manually. To ensure correctness, recent methods~\cite{selfrefine2023madaan, chen2023teaching} commonly generate multiple synthetic reasoning paths and select those that yield the correct final answer. This answer-based filtering mechanism provides a scalable proxy for ground truth supervision, yet it suffers from a critical limitation: it implicitly assumes that a correct answer is indicative of a valid reasoning process. In practice, this assumption frequently breaks down, as models may arrive at the right answer via spurious, shortcut-driven, or logically invalid intermediate steps~\cite{zhang2022automatic, bai2022training}. Such flawed reasoning paths, though superficially accurate, can propagate latent logical inconsistencies and reinforce brittle heuristics within the model.

Several efforts have been made to improve the quality of synthetic reasoning data through more rigorous verification. Code-based checking has been successfully applied in mathematical and code domains, where reasoning steps can be translated into executable programs~\cite{lewkowycz2022solving, patil2023pythia}. Yet this technique falls short in natural language reasoning tasks that involve abstract semantics, ambiguous phrasing, or commonsense knowledge. Alternatively, symbolic reasoning engines offer a promising route for validating logical consistency by translating natural language into formal logic representations~\cite{creswell2022selection, liang2021lesson,jiang2024leanreasoner}. Despite their precision, they depend heavily on sufficiently complete and unambiguous context, making them brittle in open-domain or under-specified scenarios. For natural language reasoning tasks that do not have complete and standardized context, there is currently no good method for intermediate process verification of synthetic data.

In this work, we focus on improving the quality of synthetic reasoning data under context-limited natural language reasoning settings, where traditional verification methods struggle. Inspired by classical syllogistic reasoning~\cite{wan2025srfotsyllogisticreasoningframeworkthought, constant2024bayesian}, we propose a novel data generation framework that combines the generative power of LLMs with the precision of symbolic logic engines. LLMs possess rich world knowledge and the ability to actively identify reasoning directions, which provide the necessary contextual information for the engine. Meanwhile, the reasoning engine performs automated inference based on given premises and rules, ensuring logical rigor~\cite{rabe2020mathematical, liang2017neural}. We aim to leverage the complementary strengths of LLMs and reasoning engines through fine-grained interactions to generate high-quality reasoning data.

Specifically, in this work, we propose ORACLE, a structured reasoning data generation framework that combines template-guided step-by-step generation with symbolic supervision to enable both interpretability and verifiability across multi-step chains of thought. ORACLE employs a unified response template that guides LLMs in discrete, modular steps—each comprising a clearly defined query, relevant facts, applied rules, and a revision process. These structured outputs not only facilitate automatic extraction and fine-grained analysis but also enable seamless integration with symbolic reasoning engines as external verifiers.

Our ORACLE consists of a two-stage training process. In the first stage, we use a small set of manually crafted demonstrations to bootstrap the generation of large-scale structured reasoning data via few-shot prompting. After rigorous filtering, this data is used to fine-tune the base model to internalize the step-wise reasoning format. In the second stage, we combine reasoning engine verification and large model evaluation to generate supervised fine-tuning data and preference data through a beam search strategy.

We conducted comprehensive experiments on six diverse reasoning datasets, covering symbolic, factual and commonsense tasks, using LLaMA, Mistral, and Qwen~\cite{touvron2023llama, jiang2023mistral7b, bai2023qwen}. Our ORACLE consistently achieves the best or near-best performance across all datasets and model variants, surpassing strong baseline methods. These results underscore the effectiveness and broad applicability of our ORACLE in enhancing the reasoning capabilities of large language models.

\section{Related Work}
\paragraph{Synthetic Reasoning Data Generation.}
To overcome the scarcity of high-quality reasoning data, numerous studies have explored leveraging large language models (LLMs) themselves to generate synthetic datasets~\cite{zelikman2022star, yuan2023scaling, xu2023selfrewarding}. A common strategy involves self-augmented prompting, where models generate chain-of-thought (CoT) explanations alongside answers, followed by filtering based on answer correctness. While effective in scaling up data, this method often retains reasoning paths that are spuriously correct, leading to models that memorize artifacts instead of acquiring genuine reasoning capabilities~\cite{turpin2023language,zelikman2022star}. Our ORACLE addresses this issue by supervising the entire reasoning chain, not just the final answer, thereby mitigating shortcut learning.

\paragraph{Verification and Programmatic Supervision.}
Existing approaches to improving the quality of synthetic reasoning data often rely on programmatic supervision to verify intermediate steps. In mathematical and algorithmic domains, code-based execution offers an effective means of validation by aligning reasoning steps with executable semantics~\cite{lewkowycz2022solving, patil2023pythia, leang2025theorem}. However, such methods are inherently domain-specific and fall short when applied to natural language reasoning tasks involving abstract concepts, ambiguous expressions, or commonsense knowledge. Alternatively, symbolic reasoning engines have been explored to assess logical consistency by converting natural language statements into formal logic representations~\cite{creswell2022selection, jiang2024leanreasoner, kamoi2024can}. Despite their precision, these systems are fragile in practice, often requiring complete and unambiguous context that is difficult to obtain in open-domain scenarios. In contrast, ORACLE incorporates symbolic supervision in a soft and LLM-compatible manner by structuring reasoning into discrete, interpretable modules. This hybrid design facilitates automatic verification while preserving adaptability to diverse reasoning domains, thereby addressing key limitations of existing verification techniques.

\paragraph{LLM Supervision and Preference Training.}
Fine-tuning LLMs with high-quality, structured supervision has proven effective in aligning model outputs with human reasoning patterns~\cite{ouyang2022training,rafailov2023direct}. Beyond supervised fine-tuning, preference training via comparison data~\cite{ziegler2019fine} has emerged as a powerful tool for refining generation quality. Our ORACLE first applies supervised fine-tuning (SFT) on verified reasoning paths to ground the model in faithful intermediate steps, and then leverages symbolic validation scores to construct preference pairs for Direct Preference Optimization (DPO), enabling alignment not only with correct outcomes but also with faithful reasoning.

\section{Methodology}

\begin{figure}[t]
    \centering
    \includegraphics[width=0.6\linewidth]{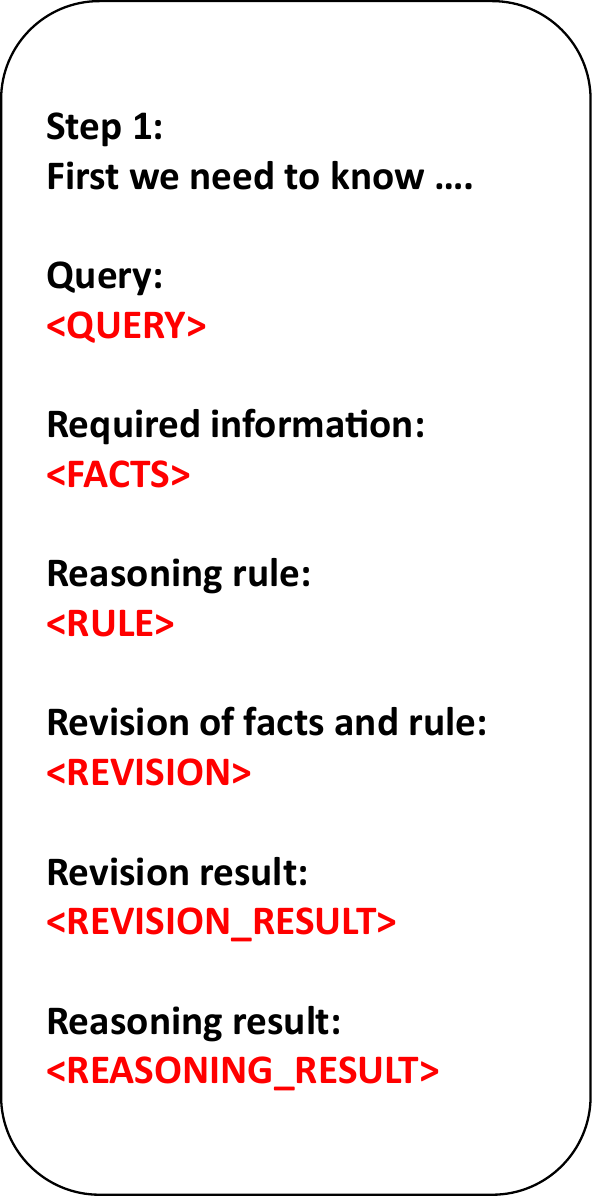}
    \caption{An overview of our structured reasoning template used during data generation of our ORACLE. Each reasoning step consists of modular fields: \texttt{<QUERY>}, \texttt{<FACTS>}, \texttt{<RULE>}, \texttt{<REVISION>}, \texttt{<REVISION\_RESULT>}, and \texttt{<REASONING\_RESULT>}. This design promotes interpretable reasoning, enables symbolic verification, and facilitates automatic extraction via pattern matching.}
    \label{fig:template}
\end{figure}

In this section, we will explain in detail how our ORACLE works.
\subsection{Template Design}

To facilitate structured reasoning and enable effective information extraction, our ORACLE design a fixed-format response template. This template guides the step-by-step reasoning process of large language models (LLMs), ensuring that key reasoning components can be easily identified and extracted using simple regular expressions. Figure~\ref{fig:template} illustrates the complete structure of our template.

In our template, the \texttt{\textless QUERY\textgreater} field denotes the sub-problem that needs to be solved at the current step. The field \texttt{\textless FACTS\textgreater} contains the relevant contextual information or premises used to solve the query. The field \texttt{\textless RULE\textgreater} specifies the reasoning principle or the logical rule applied during inference. The contents of \texttt{\textless FACTS\textgreater} and \texttt{\textless RULE\textgreater} conform to the expression form of the premises and rules in a syllogism. To improve robustness and interpretability, we also introduce a field of \texttt{\textless REVISION\textgreater}, which instructs the model to reflect on the appropriateness and sufficiency of the selected facts and the rule. The outcome of this reflective process is recorded in the \texttt{\textless REVISION\_RESULT\textgreater} field, indicating whether the original input is retained or revised. Finally, the \texttt{\textless REASONING\_RESULT\textgreater} field stores the conclusion derived from applying the reasoning rule to the given facts. During Stage 2, within the beam search process, the \texttt{\textless REASONING\_RESULT\textgreater} field is populated with the output of the symbolic reasoning engine, conditioned on the successful execution of the reasoning step. The language model follows this template iteratively, filling in each field at every reasoning step until the final answer is reached. 

\begin{figure*}[t]
  \centering
  \includegraphics[width=0.8\textwidth]{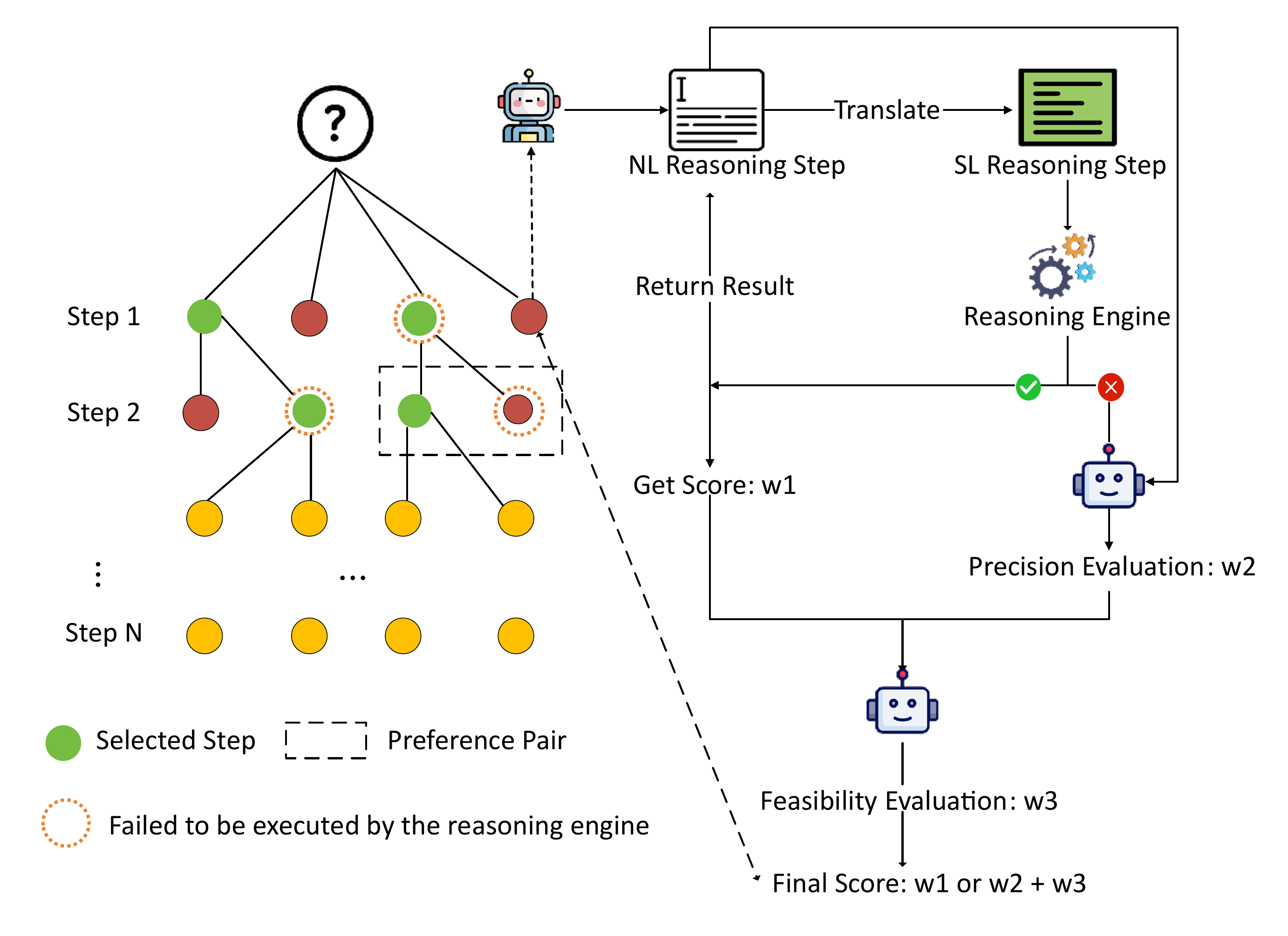}
  \caption{An overview of the beam search integrating a symbolic reasoning engine and LLM-based evaluation of our ORACLE. NL means natural language and SL means symbolic language. For each step, the LLM generates a candidate reasoning step in natural language, which is then translated into symbolic form and passed to the reasoning engine for execution. Each candidate is scored based on execution success (\(W_1\)), LLM's precision assessment (\(W_2\)), and feasibility estimation (\(W_3\)). Candidates that are successfully executed receive a final score of \(W_1 + W_3\), while others are scored as \(W_2 + W_3\). The top-\(K\) candidates based on these scores are selected and expanded in the next layer. Complete reasoning paths that produce correct final answers are collected as supervised fine-tuning(SFT) data. Additionally, preference pairs are constructed by comparing symbolically validated nodes with their invalid siblings for Direct Preference Optimization(DPO).}
  \label{fig:beamsearch_process}
\end{figure*}

\subsection{Model Training}

Our ORACLE divide the model training process into two stages, as illustrated in Figure~\ref{fig:training_pipeline}. In the first stage, we use a small number of manually crafted reasoning examples formatted with our predefined template to prompt the base LLM in a few-shot manner. This process generates a larger set of synthetic reasoning samples. We then apply a strict filtering mechanism, retaining only those samples that strictly conform to the template structure and yield correct final answers. The filtered data is used to fine-tune the base LLM. The objective of this stage is to encourage the model to internalize the step-by-step reasoning format imposed by the template and to produce structured outputs accordingly.

In the second stage, we integrate the fine-tuned LLM with a reasoning engine. We conduct multiple rounds of interaction between the model and the reasoning engine to generate two types of data: supervised fine-tuning data and preference data, based on the beam search strategy. These are used in succession to further train the base LLM via supervised fine-tuning(SFT)~\cite{ouyang2022training} and Direct Preference Optimization (DPO)~\cite{rafailov2023direct}. The goal of this stage is to use the reasoning engine as an external verifier to guide the generation of high-quality reasoning data.

\begin{table*}
  \centering
  \scriptsize
  % \resizebox{\textwidth}{!}{
  % \setlength{\tabcolsep}{2pt}
  \begin{tabular}{c|c|cccccc}
    \hline
      Models & Methods & \textbf{ProntoQA} & \textbf{ProofWriter} & \textbf{BoolQ} & \textbf{CosmosQA} & \textbf{ScienceQA} & \textbf{StrategyQA} \\
      \hline
   Llama-3.1-8B-Instruct & CoT(zero-shot)~\cite{wei2022chain} & 94.3 & 53.7 &  85.0 & 74.2 & 90.5 & 89.9 \\
    & CoT(four-shot)~\cite{wei2022chain} & 92.4 & 60.3 &  83.2 & 79.5 & 91.9 & 88.6 \\
    & RFT~\cite{yuan2023scaling} & 95.0 & 62.0 &  86.3 & 80.4 & \textbf{92.4} & 89.9 \\
     & ToT-SFT~\cite{yao2023tree} & 93.6 & 57.0 &  82.5 & 74.7 & 90.7 & 85.2 \\
      & self-rewarding~\cite{xu2023selfrewarding} & 95.8 & 63.4 &  86.2 & 81.5 & 92.2 & 90.8 \\
       & ORACLE(Ours) & \textbf{97.2} & \textbf{67.5} & \textbf{87.7} & \textbf{82.4} & 91.7 & \textbf{91.8} \\
    \hline
   Mistral-7B-Instruct-v0.3 & CoT(zero-shot)~\cite{wei2022chain} & 49.3 & 36.3 &  83.1 & 70.6 & 75.3 & 87.3 \\
    & CoT(four-shot)~\cite{wei2022chain} & 66.3 & 50.5 &  80.5 & 71.1 & 77.0 & 88.1 \\
    & RFT~\cite{yuan2023scaling} & 81.3 & 53.6 &  84.4 & 74.0 & 81.0 & 89.0 \\
     & ToT-SFT~\cite{yao2023tree} & 82.8 & 54.8 &  80.9 & 72.4 & 77.4 & 84.4 \\
      & self-rewarding~\cite{xu2023selfrewarding} & 81.6 & 58.8 &  85.6 & 71.7 & 80.2 & 89.0 \\
       & ORACLE(Ours) & \textbf{87.6} & \textbf{64.5} &  \textbf{86.8} & \textbf{78.5} & \textbf{83.3} & \textbf{89.9} \\
    \hline
    Qwen-2.5-7B-instruct & CoT(zero-shot)~\cite{wei2022chain} & 97.0 & 60.0 &  82.8 & 75.9 & 82.6 & 86.9 \\
    & CoT(four-shot)~\cite{wei2022chain} & 98.0 & 63.8 &  83.0 & 77.1 & 84.9 & 89.9 \\
    & RFT~\cite{yuan2023scaling} & 98.5 & 63.5 &  84.7 & 76.0 & 86.3 & 87.1 \\
     & ToT-SFT~\cite{yao2023tree} & 96.7 & 54.8 & 82.1 & 73.9 & 82.9 & 88.2 \\
      & self-rewarding~\cite{xu2023selfrewarding} & \textbf{98.8} & 64.5 &  83.8 & 78.1 & 82.5 & 91.5 \\
       & ORACLE(Ours) & 97.7 & \textbf{68.2} &  \textbf{87.5} & \textbf{82.1} & \textbf{87.3} & \textbf{92.6} \\
    \hline
    
  \end{tabular}
  % }
  \caption{\label{tab: main-results}
    Comparison of various reasoning and data generation methods across six datasets and three models. Our proposed method consistently achieves the best or near-best performance on all datasets and models, demonstrating its effectiveness across symbolic, factual, and commonsense reasoning tasks.
  }
\end{table*}

\subsection{Beam Search with LLM Evaluation and Reasoning Engine Validation}

To guide the generation of high-quality reasoning paths and construct effective supervision and preference signals, our ORACLE incorporate a beam search strategy that integrates both the large language model and a symbolic reasoning engine, combining the generation ability of a large language model and the rigor of the reasoning engine.

Specifically, the generation of each node in beam search involves the following process. Given an input question $q$ and a sequence of previously generated reasoning steps $x_1, x_2, \ldots$, the LLM fine-tuned in stage 1 will produce the next natural language reasoning step $x_i$. This step is then translated into a symbolic form by the base LLM with a designed prompt and passed to the reasoning engine. If the symbolic expression is executed successfully, the step receives a score $W_1$, and the result of the execution is integrated back into the natural language step. otherwise, the LLM performs a precision evaluation of $x_i$ to produce a score $W_2$. Additionally, regardless of execution success, the LLM evaluates the feasibility of $x_i$, assigning a score $W_3$. The final score for the step is computed as $W_1 + W_3$ if it is successfully executed, or $W_2 + W_3$ otherwise.

Assume the beam width at each layer is $W$, and we select the top $K$ candidates with the highest scores to expand at the next step. Each selected node is allowed to generate $W / K$ children, ensuring that the overall beam width remains constant. Among the resulting reasoning paths, we retain only the complete paths that reach a final answer and produce a correct answer. These paths are then used as supervised fine-tuning data.

To construct preference pairs for DPO training, we backtrack each reasoning path that successfully leads to the correct final answer and identify intermediate steps (nodes) that can be successfully executed and verified by the symbolic reasoning engine. For each such validated node \(x_i\), if there exists a sibling node \(x_j\) that fails engine verification, we construct a preference pair \((x_i, x_j)\), indicating that \(x_i\) is preferred over \(x_j\). 

This process is illustrated in Figure~\ref{fig:beamsearch_process}, which shows how symbolic reasoning engine validation and language model evaluation are combined within the beam search.

\begin{table*}
  \centering
  \scriptsize
  % \resizebox{\textwidth}{!}{
  \begin{tabular}{c|c|cccccc}
    \hline
      Models & Methods & \textbf{ProntoQA} & \textbf{ProofWriter} & \textbf{BoolQ} & \textbf{CosmosQA} & \textbf{ScienceQA} & \textbf{StrategyQA} \\
      \hline
   Llama-3.1-8B-Instruct & w/o engine & 96.1 & 67.0 & \textbf{87.9} & 81.9 & 90.5 & 89.9 \\
    & w/o beam search & 96.5 & 66.7 & 87.6 & 82.2 & 91.2 & 90.5 \\
    & w/o DPO  & 96.6 & 67.2 & 87.6 & 82.0 & 91.4 & 91.1 \\ 
   & ORACLE(Ours) & \textbf{97.2} & \textbf{67.5} & 87.7 & \textbf{82.4} & \textbf{91.7} & \textbf{91.8} \\
    \hline
   Mistral-7B-Instruct-v0.3 & w/o engine & \textbf{88.0} & 63.8 & 86.1 & 78.3 & 82.9 & 88.8 \\
    & w/o beam search & 86.9 & 64.2 & 86.6 & \textbf{79.1} & 83.2 & 89.4 \\
    & w/o DPO  & 87.2 & 64.4 & 86.4 & 78.8 & 83.1 & 89.6 \\ 
       & ORACLE(Ours) & 87.6 & \textbf{64.5} &  \textbf{86.8} & 78.5 & \textbf{83.3} & \textbf{89.9} \\
    \hline
    Qwen-2.5-7B-instruct & w/o engine & 97.2 & 68.0 & 86.7 & \textbf{82.4} & 86.8 & 91.5 \\
    & w/o beam search & 97.5 & 68.3 & 87.2 & 82.1 & 86.9 & 92.5 \\
    & w/o DPO  & 97.4 & \textbf{68.6} & 87.4 & 82.2 & 87.1 & 92.3 \\ 
       & ORACLE(Ours) & \textbf{97.7} & 68.2 &  \textbf{87.5} & 82.1 & \textbf{87.3} & \textbf{92.6} \\
    \hline
    
  \end{tabular}
  % }
  \caption{\label{tab:ablation-results}Ablation study on key components of our framework. We compare our full method with three ablated variants to analyze the contribution of key components: w/o engine removes reasoning engine's guidance during data generation; w/o beam search disables beam-based search while retaining the reasoning engine; w/o DPO omits Direct Preference Optimization and saves other components during training. While our full method does not always yield the best score on every dataset-model pair, it consistently achieves strong performance and often outperforms ablations across most settings, validating the effectiveness of combining symbolic guidance, beam search, and preference learning.
  }
\end{table*}

\section{Experiments}

\subsection{Experiment Setup}
\subsubsection{Datasets.} 

We evaluate our ORACLE on six datasets covering diverse types of reasoning: ProntoQA~\cite{jin2023prontoqa}, ProofWriter~\cite{talmor2020proofwriter}, BoolQ~\cite{clark2019boolq}, CosmosQA\cite{huang2019cosmos}, ScienceQA\cite{lu2022learn}, and StrategyQA\cite{geva2021strategyqa}. These datasets span multiple reasoning paradigms, including logical reasoning, factual reasoning, commonsense reasoning, and scientific multi-hop reasoning, allowing for comprehensive and effective evaluation of the reasoning capabilities of LLMs. Although ProntoQA and ProofWriter have standardized contexts, they are representative datasets for verifying logical reasoning ability. We hope to verify that our ORACLE can also achieve better results on these datasets.

\subsubsection{Baselines.} We compare our method with the following baselines: (1) \textbf{CoT}~\cite{wei2022chain}, where the model is prompted to generate step-by-step reasoning using chain-of-thought prompting without fine-tuned. We evaluate both zero-shot and four-shot settings with greedy decoding. (2) \textbf{RFT}~\cite{yuan2023scaling}, where reasoning paths are generated via CoT prompting and filtered using rejection sampling to retain only those that lead to correct answers. (3) \textbf{ToT-SFT}~\cite{yao2023tree}, where multiple reasoning paths are explored using Tree-of-Thoughts, and those with correct final answers are selected as training data for supervised fine-tuning. In our work, we use the prompt in github~\cite{tree-of-thought-prompting} to generate. (4) \textbf{Self-rewarding}~\cite{xu2023selfrewarding}, where multiple reasoning paths are generated for each question via CoT prompting, and paths with correct answers are used for supervised fine-tuning. Additionally, the LLM is prompted to score these paths, and the highest- and lowest-scoring answers for the same question are used to construct preference pairs for DPO training.

\subsubsection{Implementation details.} Our experiments are conducted using the instruction-tuned variants of widely adopted open-source LLMs, including LLaMA-3.1-8B-Instruct, Mistral-7B-Instruct-v0.3, and Qwen-2.5-7B-Instruct. For efficient training, we apply Low-Rank Adaptation (LoRA)~\cite{hu2022lora} with a rank of 8 to all models. When generating data, for all baselines other than CoT, we use four-shot prompting and sample six completions per question. For our ORACLE, the data generation in the first stage uses two-shot prompting. In the second stage, we employ a beam search strategy with a beam width $w=9$, selecting $k=3$ top-scoring nodes at each step, each of which generates $w/k=3$ children. During evaluation, all methods are applied in a zero-shot setting except for CoT. We use Pyke~\cite{frederiksen2008applying} as the symbolic reasoning engine. For scoring during beam search, if the symbolic reasoning step is successfully executed by the reasoning engine, it receives a score of $w_1=3$; otherwise, it is evaluated by the LLM for correctness, receiving $w_2=2$ if passed, or $w_2=0$ otherwise. In all cases, the LLM also conducts a feasibility evaluation, assigning $w_3=5$ if passed, and $w_3=0$ otherwise. The final score is the sum of $w_1$ or $w_2$ and $w_3$.
During data generation, we use a temperature of 1.0, while for inference, the temperature is set to 0.01. The learning rates for SFT and DPO training are $5 \times 10^{-6}$ and $1 \times 10^{-4}$, respectively. For all methods that use SFT or DPO, SFT used 12k samples; DPO used 2k preference pairs after generation and filtering. We use a batch size of 16 (with gradient accumulation) and optimize the model using AdamW. All experiments are conducted on NVIDIA A100 GPUs. Models are trained for 3 epochs. 

\subsection{Experimental Results and Analyses}

\subsubsection{Performance on ScienceQA.} ScienceQA requires complex scientific and multi-hop reasoning. Our ORACLE achieves strong and stable performance across three LLMs, reaching 91.7\%, 83.3\%, and 87.3\% accuracy on LLaMA-3.1-8B-Instruct, Mistral-7B-Instruct-v0.3, and Qwen-2.5-7B-Instruct respectively. These results consistently surpass most baselines and demonstrate the effectiveness of our approach in enhancing scientific reasoning ability.

\subsubsection{Performance on ProntoQA and ProofWriter.} On the symbolic reasoning datasets ProntoQA and ProofWriter, our ORACLE significantly outperforms all compared methods. For ProntoQA, it improves accuracy by 1.4\% to 6.0\% across models, achieving up to 97.7\%. On ProofWriter, the gains are even more pronounced, with increases of 3.7\% to 5.7\%, reaching a peak accuracy of 68.2\%. These substantial improvements highlight the superiority of our method in handling complex logical inference tasks.

\subsubsection{Performance on StrategyQA, BoolQ and CosmosQA.} For factual verification and commonsense reasoning, our ORACLE leads with notable margins. It achieves up to 91.8\%, 87.7\%, and 82.4\% on LLaMA-3.1-8B-Instruct, 89.9\%, 86.8\%, and 78.5\% on Mistral-7B-Instruct-v0.3, and 92.6\%, 87.5\%, and 82.1\% on Qwen-2.5-7B-Instruct for StrategyQA, BoolQ, and CosmosQA respectively. These improvements over the baselines, ranging from 0.9\% to 6.1\%, highlight the effectiveness of our approach in enhancing performance across a wide spectrum of reasoning tasks. The consistent gains demonstrate robust generalization capabilities, indicating that ORACLE can adapt well to diverse reasoning challenges with varying levels of complexity.

Overall, our ORACLE achieves consistent and significant performance gains across six diverse datasets and three models. This validates the effectiveness and generalization ability of our ORACLE in enhancing both logical reasoning, factual reasoning, commonsense reasoning, and scientific multi-hop reasoning capabilities.

\subsection{Ablation Study}

\begin{table*}
  \centering
  \scriptsize
  % \resizebox{\textwidth}{!}{
  \begin{tabular}{c|c|cccccc}
    \hline
      Models & Iterations & \textbf{ProntoQA} & \textbf{ProofWriter} & \textbf{BoolQ} & \textbf{CosmosQA} & \textbf{ScienceQA} & \textbf{StrategyQA} \\
      \hline
   Llama-3.1-8B-Instruct & 1st iteration & 82.7 & 52.3 & 45.2 & 55.2 & 30.3 & 24.4 \\
   & 2nd iteration & 84.1 & 55.7 & 46.3 & 56.1 & 32.1 & 25.7 \\
   \hline
   Mistral-7B-Instruct-v0.3 & 1st iteration & 64.1 & 45.2 & 40.6 & 43.7 & 31.8 & 20.3 \\
    & 2nd iteration & 66.3 & 46.1 & 41.2 & 44.0 & 32.5 & 21.0 \\
    \hline
    Qwen-2.5-7B-instruct     & 1st iteration & 57.5 & 48.7 & 33.1 & 47.8 & 25.5 & 32.8 \\
    & 2nd iteration & 58.4 & 49.2 & 33.0 & 48.5 & 25.3 & 33.4 \\

    \hline
    
  \end{tabular}
  % }
  \caption{\label{tab: success-rate}
   The success rate of the reasoning steps executed by the reasoning engine in our ORACLE on six datasets. With more training iterations, the reasoning engine is able to successfully execute a greater proportion of reasoning steps.
  }
\end{table*}

\begin{table*}
  \centering
  \scriptsize
  \begin{tabular}{c|c|cccccc}
    \hline
      Models & Error Reasons & \textbf{ProntoQA} & \textbf{ProofWriter} & \textbf{BoolQ} & \textbf{CosmosQA} & \textbf{ScienceQA} & \textbf{StrategyQA} \\
      \hline
   Llama-3.1-8B-Instruct & Generation Error & 24.0 & 40.0 & 44.0 & 38.0 & 72.0 & 80.0 \\
   & Translation Error & 76.0 & 60.0 & 56.0 & 62.0 & 28.0 & 20.0 \\
   \hline
  \end{tabular}

  \caption{\label{tab: error-distribution}
   Error type proportions of our ORACLE on Llama-3.1-8B-Instruct across six datasets when the reasoning engine fails to execute. Generation Error means that the proposed facts or rules are overly complex or incorrectly formatted to be effectively translated into symbolic language. Translation Error means that the limited translation capability of the LLM leads to syntactic or semantic errors in the symbolic representation.
  }
\end{table*}

We perform an ablation study to assess the contributions of three key components in our ORACLE: the reasoning engine, beam search, and Direct Preference Optimization (DPO). Results across six datasets and the three models are shown in Table~\ref{tab:ablation-results}.

\subsubsection{Reasoning Engine.} Removing the symbolic reasoning engine leads to the most significant performance drops, especially on tasks involving multi-step reasoning such as ProofWriter and StrategyQA. For instance, on LLaMA-3.1-8B-Instruct, accuracy on StrategyQA declines from 91.8\% to 89.9\%, confirming the engine's role in enforcing logical consistency.

\subsubsection{Beam Search.} Disabling beam search results in moderate but consistent decreases across most tasks and models. This indicates that search-based generation enhances the model's ability to explore diverse and accurate reasoning paths, particularly when combined with symbolic validation.

\subsubsection{Direct Preference Optimization.} Removing the DPO stage results in slight but systematic declines in performance, with more pronounced effects observed on tasks that demand subtle semantic interpretation and contextual alignment, such as BoolQ and ScienceQA. These tasks often require models to distinguish between superficially plausible and truly correct reasoning chains, a capability that is significantly enhanced by preference-based fine-tuning. By explicitly aligning model outputs with human-preferred reasoning patterns, DPO contributes to improving not only the final answer correctness but also the faithfulness and interpretability of intermediate steps. 

Overall, our full system achieves the best or near-best results in almost all settings, highlighting the complementary benefits of symbolic guidance, search diversity, and preference alignment. We therefore retain all components in the final framework.

\subsection{Verification of Reasoning Engine}

We further investigate the verification capability of the reasoning engine within our ORACLE. Specifically, we analyze the interaction between the reasoning engine and 1,000 reasoning steps generated by each of models across six datasets. For each case, we record the success rate with which the reasoning engine executes the generated steps.

Table~\ref{tab: success-rate} presents the success rates of reasoning steps that are successfully executed by the symbolic reasoning engine across six datasets. Notably, tasks with a stronger emphasis on symbolic reasoning, such as ProofWriter and ProntoQA, exhibit significantly higher success rates, especially for Llama-3.1-8B-Instruct, reaching 52.3\% and 82.7\% respectively. This indicates that our template-driven framework effectively guides LLMs to generate reasoning steps that align well with formal symbolic logic, facilitating correct execution by the engine.

In contrast, datasets involving more diverse or commonsense reasoning, such as BoolQ, ScienceQA, and StrategyQA, show relatively lower success rates across all models. These tasks place higher demands on the model's inherent reasoning and translation abilities. The LLMs are required not only to generate appropriate premises and rules that conform to the norms of deductive reasoning but also to accurately translate natural language into the corresponding symbolic language.

However, as the numbers of iterations increase, the proportion of reasoning steps successfully executed by the reasoning engine also increases, indicating that through training, LLM not only enhances its reasoning ability, but also enhances its ability to generate standardized reasoning steps.

To analyze the reasons for the failure of the reasoning engine, we analyzed 50 failure examples of our ORACLE on Llama-3.1-8B-Instruct across each of six datasets, and finally summarized them into two main reasons: i) the proposed facts or rules are overly complex or ill-formed to be effectively translated into symbolic language; and ii) the model's limited translation capabilities resulted in syntactic or semantic errors in the symbolic representation. These are also future improvements we will pursue in ORACLE. Specific error type proportions can be found in Table~\ref{tab: error-distribution}.

Although the reasoning engine does not achieve a high execution success rate on certain tasks, our ORACLE enhances robustness by additionally leveraging the LLM to assess the correctness and feasibility of reasoning. The effectiveness of our ORACLE is further validated by the experimental results presented in Table~\ref{tab: main-results}.

\section{Conclusion}

In this work, we present ORACLE, a structured synthetic reasoning data generation framework that integrates fixed-format prompting, symbolic reasoning supervision, and beam search-based data selection. ORACLE focuses on generating high-quality, verifiable multi-step reasoning data for LLMs tuning to improve reasoning performance. Experimental results across multiple models and reasoning datasets confirm that training on ORACLE-generated data leads to significant improvements in reasoning accuracy. In future work, we plan to explore broader reasoning paradigms and verification strategies to further enhance the quality and applicability of synthetic supervision.

\bibliography{aaai2026}

\end{document}